\documentclass[]{interact}

\usepackage{epstopdf}
\usepackage[caption=false]{subfig}
\usepackage{mathtools}
\usepackage{fix-cm}
\usepackage{amsmath}

\usepackage[square,numbers]{natbib}
\renewcommand\bibfont{\fontsize{10}{12}\selectfont}

\theoremstyle{plain}
\newtheorem{theorem}{Theorem}[section]
\newtheorem{lemma}[theorem]{Lemma}

\newtheorem{proposition}[theorem]{Proposition}

\theoremstyle{definition}

\theoremstyle{remark}
\newtheorem{remark}{Remark}

\DeclareMathOperator*{\argmin}{arg\,min}

\begin{document}


\title{Higher-Order Asymptotics of Test-Time Adaptation for \\ Batch Normalization Statistics}

\author{
\name{Masanari Kimura\thanks{CONTACT Masanari Kimura. Email: m.kimura@unimelb.edu.au}}
\affil{School of Mathematics and Statistics, The University of Melbourne}
}

\maketitle

\begin{abstract}
This study develops a higher‐order asymptotic framework for test‐time adaptation (TTA) of Batch Normalization (BN) statistics under distribution shift by integrating classical Edgeworth expansion and saddlepoint approximation techniques with a novel one‐step M‐estimation perspective.
By analyzing the statistical discrepancy between training and test distributions, we derive an Edgeworth expansion for the normalized difference in BN means and obtain an optimal weighting parameter that minimizes the mean-squared error of the adapted statistic. 
Reinterpreting BN TTA as a one‐step M‐estimator allows us to derive higher‐order local asymptotic normality results, which incorporate skewness and other higher moments into the estimator’s behavior.
Moreover, we quantify the trade-offs among bias, variance, and skewness in the adaptation process and establish a corresponding generalization bound on the model risk.
The refined saddlepoint approximations further deliver uniformly accurate density and tail probability estimates for the BN TTA statistic.
These theoretical insights provide a comprehensive understanding of how higher-order corrections and robust one‐step updating can enhance the reliability and performance of BN layers in adapting to changing data distributions.
\end{abstract}


\section{Introduction}
Batch Normalization (BN)~\citep{ioffe2015batch,bjorck2018understanding,santurkar2018does} has become a fundamental component in modern neural network architectures, significantly accelerating training and improving generalization by mitigating internal covariate shift.
Traditionally, BN layers rely on the statistical estimates computed from training data during both training and inference.
However, when models are deployed in environments where the test data distribution deviates from the training distribution, a scenario commonly referred to as distribution shift~\citep{shimodaira2000improving,perry2005climate,moreno2012unifying,quinonero2022dataset,kimura2022information,kimura2024short}, the fixed training statistics may no longer be representative.
This mismatch can degrade model performance, motivating the development of test-time adaptation (TTA) strategies~\citep{kimura2021understanding,he2021autoencoder,chen2022contrastive,kimura2024test,kim2024test,liang2025comprehensive} that update BN statistics using incoming test data \citep{li2018adaptive,yang2022test,lim2023ttn}.
Despite promising empirical results, many existing TTA methods for BN are primarily heuristic.
To address this challenge, a rigorous and comprehensive statistical analysis is needed.
Thus, we aim to not only characterize the asymptotic behavior of BN statistics but also guide the design of optimal adaptation strategies.
In this study, we develop a higher-order asymptotic framework for BN TTA.
Our analysis leverages the classical Edgeworth expansion to approximate the distribution of the normalized difference between training and test BN means, thereby capturing the influence of higher-order moments.
This expansion enables us to derive an optimal blending parameter that minimizes the mean-squared error of the adapted statistic.
By quantitatively balancing the contributions of bias, variance, and skewness, our framework provides clear insights into when and how much the test BN statistics should be trusted relative to the training estimates.
Furthermore, we extend our theoretical analysis using saddlepoint approximation techniques to obtain uniformly accurate estimates of the density and tail probabilities for the BN TTA statistic.
These refined approximations not only offer improved finite sample corrections but also support the derivation of a generalization bound on the model risk under distribution shift.

\subsection{Practical Relevance of Theoretical Results}
The theoretical findings in this study have novel implications for the practical application of test-time adaptation in batch normalization.
The derivation of the optimal weighting parameter in Proposition~\ref{prp:optimal_lambda} provides a clear guideline on how to balance the contributions of training and test BN statistics.
Moreover, Theorem~\ref{thm:generalization_bound} establishes a formal bound on the risk of the adapted model, demonstrating the trade-offs among bias, variance, and higher-order corrections.
The results from the saddlepoint approximation in Proposition~\ref{prp:saddle_point_density_tnm} further refine the statistical estimation of BN statistics by improving density and tail probability estimates, making adaptation more reliable even in small-batch settings.
Additionally, the reformulation of BN TTA as a one-step M-estimator in Theorem~\ref{thm:lan_one_step} provides a more robust statistical framework for dynamically adjusting BN statistics under distribution shift.
These theoretical insights contribute directly to designing more stable and effective adaptation strategies, ensuring improved generalization across different test environments

\section{Preliminary}
Let $X_1,X_2,\dots,X_n \in \mathbb{R}^d$ be i.i.d. real‐valued random variables with distribution $P$ and $Y_1,Y_2,\dots,Y_m$ be i.i.d. real‐valued random variables with distribution $Q$.
Let $\mu_P = \mathbb{E}_P[X]$, $\mu_Q = \mathbb{E}_Q[Y]$, $\Sigma_P = \text{Var}_P(X)$, $\Sigma_Q = \text{Var}_Q(Y)$, and assume that higher-order cumulants exist and are finite.
Denote the training and test BN statistics by
\begin{align}
    \hat{\mu}_{P,n} &= \frac{1}{n}\sum^n_{i=1} X_i, \quad \sigma_{P,n} = \frac{1}{n}\sum^n_{i=1}(X_i - \hat{\mu}_{P,n})^2, \\
    \hat{\mu}_{Q,m} &= \frac{1}{m}\sum^m_{i=1} Y_i, \quad \sigma_{Q,m} = \frac{1}{m}\sum^m_{i=1}(Y_i - \hat{\mu}_{Q,m})^2.
\end{align}
A typical BN layer normalizes an activation $z \in \mathbb{R}^d$ as
\begin{align}
    \text{BN}(z ; \mu, \sigma^2) = \gamma\frac{z - \mu}{\sqrt{\sigma^2 + \epsilon}} + \beta,
\end{align}
where $\gamma$ and $\beta$ are learnable affine parameters, $\mu$ and $\sigma^2$ are the mean and variance used for normalization, and $\epsilon > 0$ is a small constant for numerical stability.
Under standard inference, one uses the training estimates:
\begin{align}
    \text{BN}_{\text{train}}(z) = \gamma\frac{z - \hat{\mu}_{P,n}}{\sqrt{\hat{\sigma}^2_{P,n} + \epsilon}} + \beta,
\end{align}
In Test-Time Adaptation (TTA), the BN layer is updated using the test data statistics:
\begin{align}
    \text{BN}_{\text{TTA}}(z) = \gamma\frac{z - \hat{\mu}_{Q,m}}{\sqrt{\hat{\sigma}^2_{Q,m} + \epsilon}} + \beta.
\end{align}
Throughout the paper, we assume that both distributions $P$ and $Q$ possess moments up to the required order (at least four) to facilitate the derivation of the Edgeworth expansion and saddlepoint approximations.

\section{Main Results}
In this section, we present our theoretical contributions, beginning with an Edgeworth expansion for the BN TTA statistic and then developing saddlepoint approximations to further refine the asymptotic behavior.
These results provide insights into the roles of bias, variance, and skewness in the test-time adaptation process, as well as guidance for selecting optimal BN adaptation parameters.
\subsection{Edgeworth Expansion for BN TTA}
Our first result characterizes the asymptotic distribution of the normalized difference between the test and training BN means using an Edgeworth expansion.
The classical result of normal approximation~\citep{bhattacharya1978validity,bhattacharya2010normal} yields the following asymptotic expansion.
\begin{lemma}
    \label{lem:expansion_true_difference}
    Let the true difference in means be $\Delta\mu = \mu_Q - \mu_P$.
    Then, if we define the normalized difference as
    \begin{align}
        T_{n,m} \coloneqq \sqrt{\frac{nm}{n + m}}\left(\hat{\mu}_{Q,m} - \hat{\mu}_{P,n} - \Delta_\mu \right),
    \end{align}
    there exists an Edgeworth expansion for the c.d.f. of $T_{n,m}$ given by
    \begin{align}
        \mathbb{P}\left(T_{n,m} \leq x\right) = \Phi\left(\frac{x}{\sqrt{V_{n,m}}}\right) - \frac{\Delta_{3,n,m}}{6V_{n,m}^{3/2}}\left(\frac{x^2}{V_{n,m}} - 1\right)\phi\left(\frac{x}{\sqrt{V_{n,m}}}\right) + R_{n,m}(x),
    \end{align}
    where $\Phi$ and $\phi$ are the standard normal c.d.f. and p.d.f., respectively,
    \begin{align*}
        V_{n,m} &\coloneqq \frac{n\sigma_Q^2 + m\sigma_P^2}{n + m}, \\
        \Delta_{3,n,m} &\coloneqq \frac{\kappa_{3,Q}\alpha^3}{\sqrt{m}} - \frac{\kappa_{3,P}\beta^3}{\sqrt{n}}, \\
        \alpha &\coloneqq \sqrt{\frac{n}{n + m}},\quad \beta \coloneqq \sqrt{\frac{m}{n + m}}, \\
        \kappa_{3,P} &= \mathbb{E}\left[(X - \mu_P)^3\right],\quad \kappa_{3,Q} = \mathbb{E}\left[(Y - \mu_Q)^3\right],
    \end{align*}
    and the remainder term $R_{n,m}(x)$ is of order $O\left(\frac{1}{\sqrt{n}} + \frac{1}{\sqrt{m}}\right)$ uniformly in $x$.
\end{lemma}
The Edgeworth expansion in Lemma~\ref{lem:expansion_true_difference} provides a refined asymptotic approximation for the distribution of the normalized difference between training and test BN means.
Unlike the standard normal approximation, which considers only the first two moments, this expansion explicitly accounts for the influence of the third moment, captured by the term $\Delta_{3,n,m}$.
This term reflects the asymmetry in the underlying data distributions, which plays a crucial role in test-time adaptation.
In practical settings, if the test distribution is highly skewed, the Edgeworth correction indicates that the adapted BN statistic may systematically deviate from the normal approximation, leading to potential miscalibrations in the adaptation process.
The sign and magnitude of $\Delta_{3,n,m}$ dictate the extent of this deviation, meaning that when $\kappa_{3,Q}$ is large, direct substitution of test statistics without adjustment may introduce additional bias.

In practice, BN TTA involves updating the BN mean using a convex combination of the training and test estimates~\citep{yang2022test,lim2023ttn}:
\begin{align}
    \mu_{\text{TTA}}(\lambda) = \lambda \hat{\mu}_{P,n} + (1 - \lambda)\hat{\mu}_{Q,m}. \label{eq:running_bn}
\end{align}
Consider to quantify the error $\mu_{\text{TTA}}(\lambda) - \mu_Q$.
Then,
\begin{align*}
    \mu_{\text{TTA}}(\lambda) - \mu_Q &= \lambda \hat{\mu}_{P,n} + (1 - \lambda)\hat{\mu}_{Q,m} - \mu_Q \\
    &= \lambda(\hat{\mu}_{P,n} - \mu_Q) + (1 - \lambda)(\hat{\mu}_{Q,m} - \mu_Q) \\
    &= \lambda\left\{(\hat{\mu}_{P,n} - \mu_P) - (\mu_Q - \mu_P)\right\} + (1 - \lambda)(\hat{\mu}_{Q,m} - \mu_Q) \\
    &= \lambda\left\{(\hat{\mu}_{P,n} - \mu_P) - \Delta_\mu \right\} + (1 - \lambda)(\hat{\mu}_{Q,m} - \mu_Q).
\end{align*}
Taking the square and expectation, and assuming the errors from the two estimates are independent so that their cross term vanishes, we obtain
\begin{align}
    E(\lambda) &\coloneqq \mathbb{E}\left[(\mu_{\text{TTA}}(\lambda) - \mu_Q)^2\right] \\
    &= \lambda^2\Delta_\mu^2 + \lambda^2\text{Var}(\hat{\mu}_{P,n}) + (1 - \lambda)^2\text{Var}(\hat{\mu}_{Q,m}) + \Gamma_{P,Q,n,m} \\
    &\approx \lambda^2\Delta_\mu^2 + \frac{\hat{\sigma}_{P,n}^2}{n} + (1 - \lambda)^2\frac{\hat{\sigma}_{Q,m}^2}{m} + \Gamma_{P,Q,n,m},
\end{align}
where $\Gamma_{P,Q,n,m}$ is the skewness correction term:
\begin{align*}
    \Gamma_{P,Q,n,m} = \left|\lambda\frac{\kappa_{3,P}}{n^{3/2}} - (1 - \lambda)\frac{\kappa_{3,Q}}{m^{3/2}}\right|.
\end{align*}
We now consider to determine the optimal $\lambda^*$ that minimizes $E(\lambda)$ over $[0, 1]$.
The optimization problem is
\begin{align}
    \lambda^* = \argmin_{\lambda \in [0, 1]}\left\{\lambda^2\Delta_\mu^2 + \frac{\hat{\sigma}_{P,n}^2}{n} + (1 - \lambda)^2\frac{\hat{\sigma}_{Q,m}^2}{m} + \left|\lambda\frac{\kappa_{3,P}}{n^{3/2}} - (1 - \lambda)\frac{\kappa_{3,Q}}{m^{3/2}}\right|\right\}.
\end{align}
For analytical tractability we assume that the term inside the absolute value is nonnegative.
Differentiate $E(\lambda)$ with respect to $\lambda$ and setting the derivative equal to zero gives
\begin{align}
    2\lambda\Delta_\mu^2 + 2\lambda\frac{\hat{\sigma}_{P,n}^2}{n} - 2(1 - \lambda)\frac{\hat{\sigma}_{Q,m}^2}{m} + \frac{\kappa_{3,P}}{n^{3/2}} + \frac{\kappa_{3,Q}}{m^{3/2}} &= 0 \nonumber \\
    2\lambda\left(\Delta_\mu^2 + \frac{\hat{\sigma}_{P,n}^2}{n} + \frac{\hat{\sigma}_{Q,m}^2}{m}\right) &= \frac{2\hat{\sigma}_{Q,m}^2}{m} - \frac{\kappa_{3,P}}{n^{3/2}} - \frac{\kappa_{3,Q}}{m^{3/2}} \nonumber \\
    \lambda^* &= \frac{\frac{\hat{\sigma}_{Q,m}^2}{m} - \frac{1}{2}\left(\frac{\kappa_{3,P}}{n^{3/2}} + \frac{\kappa_{3,Q}}{m^{3/2}}\right)}{\Delta_\mu^2 + \frac{\hat{\sigma}_{P,n}^2}{n} + \frac{\hat{\sigma}_{Q,m}^2}{m}}. \label{eq:optimal_lambda}
\end{align}
This expression is the candidate minimizer, valid provided the assumed sign condition holds and that $\lambda^* \in [0, 1]$.
We can provide this result as the following statement.
\begin{proposition}
    \label{prp:optimal_lambda}
    Under the assumptions, the optimal $\lambda^*$ for the test-time adapted BN in Eq.~\eqref{eq:running_bn} is given in Eq~\eqref{eq:optimal_lambda}.
\end{proposition}
The optimal weighting parameter $\lambda^*$ in Eq.~\eqref{eq:optimal_lambda} provides a principled way to balance the contributions of the training and test BN means based on their relative reliability.
The numerator of $\lambda^*$ captures the variance and skewness adjustments, favoring the test mean more heavily when the test batch variance $\sigma_Q^2/m$ is small and when the skewness correction terms $\kappa_{3,P}/n^{3/2}$ and $\kappa_{3,Q}/m^{3/2}$ are negligible.
The denominator, which includes both the squared bias term $\Delta_\mu^2$ and the variance terms, ensures that adaptation remains well-calibrated by preventing extreme shifts when distribution mismatch is substantial.
From a practical perspective, $\lambda^*$ behaves adaptively based on the available test data.
When the test batch size $m$ is small, $\lambda^*$ remains closer to 1, meaning that the model relies more on the training BN statistics to prevent instability due to high variance in test mean estimates.
However, as $m$ grows, $\lambda^*$ gradually decreases, leading to a higher reliance on the test statistics.
This aligns with empirical observations that test-time adaptation improves with more test samples~\citep{yang2022test,lim2023ttn}. 
Moreover, the presence of skewness terms in $\lambda^*$ highlights the importance of higher-order moments in adaptation.
When $\kappa_{3,Q}$ is large, meaning the test data is highly skewed, the optimal strategy shifts towards incorporating more of the test BN mean since the training mean may be an inadequate summary of the test distribution.
This suggests that in practical applications, dynamically adjusting $\lambda^*$ over time as new test data is observed could further improve adaptation performance.
In scenarios where test distributions exhibit heavy tails or asymmetric shifts, explicitly accounting for these higher-order effects could enhance stability and generalization under distribution shift.
\begin{remark}
    With a very large test batch $m$, the optimal strategy is to rely entirely on the test statistics. That is, one should completely replace the training BN mean with the test BN mean.
    Moreover, the derivative of $\lambda^*$ with respect to $m$ is positive, meaning that as more test samples are collected the model can trust the test BN estimate more.
\end{remark}
\begin{remark}
    When the test distribution is skewed it is optimal to put more weight on the test BN mean as $m$ increases.
    In practical BN TTA implementations, this suggests that the momentum or weighting parameter should be tuned to increase as more test data becomes available.
\end{remark}
\begin{remark}
    For a model subject to distribution shift, the BN layer’s sensitivity to higher‐order moments can have an effect on performance.
    By quantitatively balancing variance and skewness, one can decide whether a model should use a frozen training BN statistic, a fully updated test BN statistic, or their linear combination.
\end{remark}
Furthermore, under additional boundedness and Lipschitz continuity assumptions, we derive a generalization bound on the risk of the BN TTA model.
\begin{theorem}
    \label{thm:generalization_bound}
    Assume that there exists a constant $B$ such that $|X_i|, |Y_j| \leq B$ almost surely and that the loss function is $L$-Lipschitz with respect to its BN–normalized outputs.
    Suppose one forms an adapted BN mean by weighting the training and test estimates as $\mu_{\text{TTA}} = \lambda^* \hat{\mu}_{P,n} + (1 - \lambda^*)\hat{\mu}_{Q,m}$ with the optimal weighting parameter chosen as in Eq.~\eqref{eq:optimal_lambda}.
    Then, with probability at least $1 - \delta$ the risk $\mathcal{R}(f_\theta)$ of the BN TTA model satisfies
    \begin{align}
        \mathcal{R}\left(f_{\theta}^{\text{TTA}}\right) &\leq \mathcal{R}\left(f_{\theta}^{\text{train}}\right) + \frac{L|\gamma|}{\sqrt{\hat{\sigma}_{P,n}^2 + \epsilon}}\Bigg\{\frac{A}{V}(|\Delta_\mu| + t_p) + \left(1 - \frac{A}{V}\right)t_Q \nonumber \\
        &\quad\quad\quad\quad\quad + \frac{1}{6V^{3/2}}\left[\frac{A}{V}\frac{|\kappa_{3,P}|}{n^{3/2}} + \left(1 - \frac{A}{V}\right)\frac{|\kappa_{3,Q}|}{m^{3/2}}\right]\Bigg\},
    \end{align}
    where
    \begin{align*}
        A &= \frac{\sigma_{Q}^2}{m} - \frac{1}{2}\left(\frac{\kappa_{3,P}}{n^{3/2}} + \frac{\kappa_{3,Q}}{m^{3/2}}\right), \quad V = \Delta_\mu^2 + \frac{\sigma_P^2}{n} + \frac{\sigma_Q^2}{m}, \\
        t_P &= \sqrt{\frac{2\sigma_P^2 \ln (4/\delta)}{n}} + \frac{2B \ln (4/\delta)}{3n}, \quad t_Q = \sqrt{\frac{2\sigma_Q^2 \ln (4/\delta)}{m}} + \frac{2B \ln (4/\delta)}{3m}.
    \end{align*}
\end{theorem}
\begin{remark}
    As $m$ increases, note that variance and skewness terms decay as $O(m^{-1})$ and $O(m^{-3/2})$, respectively.
    Consequently, $A$ becomes smaller for large $m$.
    At the same time, the term $t_Q$, which is the concentration bound for the test mean decays as $O(m^{-1/2})$.
    Therefore, in the large $m$ limit, the optimal strategy is to rely almost exclusively on the test statistic as $\lambda^* \to 0$ and the bound tightens.
\end{remark}
\begin{remark}
    The denominator $V$ aggregates the squared bias (distribution shift) and variance terms.
    A larger $V$ increases the weight in the denominator, which in turn reduces the influence of the training estimate.
\end{remark}
\begin{remark}
    The third term in the bound, which is proportional to $1 / 6V^{3/2}$, incorporates the effective skewness contributions from both the training and test distributions.
    Its impact is moderated by the factor $A / V$ for the training skewness and by $1 - A / V$ for the test skewness.
    In practice, if the test distribution exhibits substantial skewness relative to the training distribution, then more weight is given to the test BN mean, further tightening the risk bound.
\end{remark}

\subsection{Saddlepoint Approximation for BN TTA}
While the Edgeworth expansion provides a higher-order correction to the normal approximation by incorporating skewness and other moment-based adjustments, its accuracy deteriorates in the tails of the distribution.
This is particularly problematic in test-time adaptation scenarios where small batch sizes or extreme distribution shifts make accurate tail probability estimation crucial.
Edgeworth expansions, being truncated series approximations, may produce negative probability densities or fail to capture the correct tail behavior in finite-sample settings.
Saddlepoint approximations~\citep{daniels1954saddlepoint,reid1988saddlepoint,dasgupta2008saddlepoint}, on the other hand, provide a more accurate and computationally efficient alternative by leveraging the cumulant generating function (c.g.f.) to construct density and tail probability estimates.
Unlike Edgeworth expansions, saddlepoint methods do not rely on explicit moment truncation and instead use a Laplace-type integral approximation, leading to more reliable estimates even for moderate sample sizes.
As a result, saddlepoint approximations remain valid in regions where the Edgeworth expansion may fail, making them particularly valuable for test-time adaptation where the adaptation statistic must be estimated robustly under distribution shift.
Furthermore, saddlepoint approximations yield uniformly accurate density estimates across the entire range of values, ensuring that BN statistics used for adaptation are well-calibrated.
This is particularly beneficial in safety-critical applications such as medical imaging or autonomous systems, where even small misestimations in normalization statistics could lead to significant errors in model predictions.
The following results formalize the application of saddlepoint approximations to BN TTA, demonstrating their superiority in capturing the asymptotic behavior of adapted BN means under distribution shift.
\begin{lemma}
    \label{lem:cgf_tnm}
    Following the derivation for the Edgeworth expansion, the cumulant generating function $K(t) = \ln \mathbb{E}[e^{t T_{n,m}}]$ admits the following expansion.
    \begin{align}
        K(t) = \frac{1}{2}V_{n,m}t^2 + \frac{1}{6}\Delta_{3,n,m}t^3 + O(t^4). \label{eq:cgf_tnm}
    \end{align}
\end{lemma}
\begin{proposition}
    \label{prp:saddle_point_density_tnm}
    Assume that the cumulant generating function of $T_{n,m}$ is given by Eq.~\eqref{eq:cgf_tnm}
    For a given real $x$, suppose there exists a unique real solution $\hat{t} = \hat{t}(x)$ to the saddlepoint equation $K'(\hat{t}) = x$.
    Then the density $f_{T_{n,m}}(x)$ is approximated by
    \begin{align*}
        f_{T_{n,m}}(x) \approx \frac{1}{\sqrt{2\pi (V_{n,m} + \Delta_{3,n,m}\hat{t}+ O(\hat{t}^2))}}\exp\left\{\frac{V_{n,m}\hat{t}^2}{2} + \frac{\Delta_{3,n,m}\hat{t}^3}{6} - \hat{t}x + O(\hat{t}^4)\right\}.
    \end{align*}
\end{proposition}
In addition to Proposition~\ref{prp:saddle_point_density_tnm}, we now establish further results concerning the accuracy and uniform convergence of the saddlepoint approximation for the BN TTA statistic $T_{n,m}$.
By expanding $K(t)$ about the saddlepoint $\hat{t}$ and controlling the remainder in the Laplace inversion integral, one shows that the error term is bounded by a quantity proportional to $\min\{\sqrt{n},\sqrt{m}\}^{-1}$.
Then, we have the following statement.
\begin{lemma}
    \label{lem:uniform_error}
    Assume that the cumulant generating function $K(t)$ of $T_{n,m}$ is analytic in a neighborhood of the origin and that there exist constants $C > 0$ and $\xi > 0$ such that for all $|t| \leq \xi$, 
    \begin{align*}
        |K(t)| \le C|t|^2.
    \end{align*}
    Then, for any fixed $x$ satisfying $|x| \leq M$ (with $M>0$ fixed), there exists a constant $K_1>0$ (independent of $n$ and $m$) such that the relative error of the saddlepoint density approximation $\hat{f}_{T_{n,m}}(x)$ as in Proposition~\ref{prp:saddle_point_density_tnm}) satisfies
    \begin{align*}
        \left|\frac{f_{T_{n,m}}(x)-\hat{f}_{T_{n,m}}(x)}{f_{T_{n,m}}(x)}\right|
        \le \frac{K_1}{\min\{\sqrt{n},\sqrt{m}\}}.
    \end{align*}
\end{lemma}
In addition, the following result follows by applying a refined saddlepoint approximation to the tail probability via the Lugannani–Rice formula~\citep{lugannani1980saddle}).
\begin{proposition}
    \label{prp:tail_probability}
    Under the same conditions as in Lemma~\ref{lem:uniform_error}, the tail probability for $T_{n,m}$ can be accurately approximated by the Lugannani–Rice formula.
    In particular, for any real $x$, define $\hat{t} = \hat{t}(x)$ such that $K'(\hat{t}) = x$, and set
    \begin{align*}
        \hat{w} = \text{sign}(x)\sqrt{2\left(\hat{t}x-K(\hat{t})\right)}, \quad \hat{u} = \hat{t}\sqrt{K^{\prime\prime}(\hat{t})}.
    \end{align*}
    Then, the tail probability can be approximated by
    \begin{align*}
        \mathbb{P}(T_{n,m}\ge x)
        \approx 1-\Phi(\hat{w})+\phi(\hat{w})\left(\frac{1}{\hat{w}}-\frac{1}{\hat{u}}\right),
    \end{align*}
    where $\Phi$ and $\phi$ are the standard normal c.d.f. and p.d.f., respectively. Moreover, the relative error of this approximation is of order $O\Bigl(\min\{1/\sqrt{n},1/\sqrt{m}\}\Bigr)$.
\end{proposition}
By Taylor expanding $K(t)$ around the saddlepoint $\hat{t}(x)$ and applying analytic continuation, one can show that the error in the saddlepoint density approximation is uniformly small over any compact set $D$.
Uniform bounds on the remainder term, established via Lemma~\ref{lem:uniform_error}, ensure that the convergence is uniform as $\min\{n,m\}\to \infty$.
This can be given as the following statement.
\begin{theorem}
    \label{thm:uniform_convergence}
    Suppose that the random variables $X_i$ and $Y_j$ have finite moments up to order four and that the cumulant generating function $K(t)$ of $T_{n,m}$ is analytic in a neighborhood containing the saddlepoint $\hat{t}(x)$ for each $x$ in a compact set $D\subset \mathbb{R}$.
    Then, as $\min\{n,m\}\to \infty$, the saddlepoint density approximation $\hat{f}_{T_{n,m}}(x)$ converges uniformly to the true density $f_{T_{n,m}}(x)$
    \begin{align*}
        \sup_{x\in D}\left|f_{T_{n,m}}(x)-\hat{f}_{T_{n,m}}(x)\right|\to 0.
    \end{align*}
\end{theorem}
These results collectively establish a comprehensive higher-order asymptotic framework for BN test-time adaptation.
The Edgeworth expansion and the derived optimal weighting parameter provide a clear picture of how the interplay among bias, variance, and skewness affects BN statistics under distribution shift.
The saddlepoint approximations further refine this picture by delivering uniformly accurate density and tail probability estimates, which are critical for understanding the behavior of BN layers in practical deployment scenarios.

\subsection{Reformulation of BN TTA as One-Step M‐Estimator}
While the previous analyses focused on characterizing the higher-order asymptotic properties of BN TTA through Edgeworth and saddlepoint approximations, an alternative and complementary perspective can be obtained by reformulating BN TTA as a one-step M-estimator~\citep{van2000asymptotic}.
The M-estimation framework provides a principled approach to statistical estimation by solving an estimating equation rather than relying solely on moment-based approximations.
By interpreting BN TTA as an M-estimation problem, we gain access to a rich set of mathematical tools from asymptotic statistics that allow us to incorporate robust correction terms and improve the adaptation process.
One of the key advantages of this approach is that it naturally extends the adaptation mechanism beyond simple weighted averaging, enabling the derivation of one-step correction formulas that improve estimation accuracy when only a limited number of test samples are available.
Rather than treating the adaptation process as a static update rule, this framework allows BN statistics to be refined iteratively using robust estimating equations, improving stability under distribution shift.
Moreover, by analyzing the behavior of the one-step estimator under local asymptotic normality (LAN), we can derive optimal correction terms that account for higher-order moments in a systematic way.
This perspective is particularly valuable in practical scenarios where test batch sizes are small, and direct empirical estimates of test statistics may be unreliable.
By viewing BN TTA as an M-estimation problem, we can design adaptation strategies that balance robustness and efficiency, ensuring that BN updates remain stable even in challenging distribution shift settings.
In the following sections, we formally define the one-step M-estimator for BN TTA and establish its higher-order asymptotic properties.

Suppose the target parameter is the true BN mean under the test distribution $\mu_Q$.
In standard M‐estimation, one seeks $\mu$ such that
\begin{align*}
    \sum^m_{j=1}\psi(Y_j, \mu) = 0,
\end{align*}
where $\psi(y, \mu)$ is a chosen estimating function.
For a location problem, a natural starting choice is $\psi(y, \mu) = y - \mu$, which yields the classical estimator $\hat{\mu}_{Q,m}$ when solved exactly. 
However, when only a moderate number of test samples are available or when higher‐order moments such as skewness are important, one may wish to incorporate corrections via a one‐step update starting from an initial estimator.

Let the training BN mean $\hat{\mu}_{P,n}$ serve as the initial estimator.
A one‐step M‐estimator update then takes the form
\begin{align*}
    \hat{\mu}_{\text{one-step}} = \hat{\mu}_{P,n} - \frac{\sum^m_{j=1}\psi(Y_j, \hat{\mu}_{P,n})}{\sum^m_{j=1}\psi^\prime(Y_j, \hat{\mu}_{P,n})}.
\end{align*}
For the simple score function $\psi(y, \mu) = y - \mu$, the derivative $\psi^\prime(y, \mu) = -1$ and the update recovers
\begin{align*}
    \hat{\mu}_{\text{one-step}} = \hat{\mu}_{P,n} + \frac{1}{m}\sum^m_{j=1}(Y_j - \hat{\mu}_{P,n}) = \hat{\mu}_{Q,m}.
\end{align*}
Thus, a one‐step update with the quadratic loss returns the test mean.
To capture higher‐order effects, we modify the score function by incorporating, for example, a third‐order correction.
One such modification is
\begin{align*}
    \psi(y, \mu) = (y - \mu)\left[1 - \frac{1}{6}\frac{\kappa_{3,Q}}{\sigma_Q^3}(y - \mu)\right].
\end{align*}
This formulation effectively damps the correction from the test sample mean based on the third moment of the test data.
\begin{lemma}
    \label{lem:expansion_empirical_score}
    Let $Y_1,\dots,Y_m$ be i.i.d. observations from a distribution $Q$ with true parameter $\mu_0$.
    Let $\psi(\mu, y)$ be a function that is three times continuously differentiable in $\mu$ in a neighborhood of $\mu_0$ and suppose that $\mathbb{E}[\psi(Y, \mu_0)] = 0$.
    Define
    \begin{align*}
        \psi^\prime_0 \coloneqq -\mathbb{E}\left[\frac{\partial}{\partial\mu}\psi(Y, \mu_0)\right],\quad \text{and} \quad \psi^{\prime\prime}_0 \coloneqq \mathbb{E}\left[\frac{\partial^2}{\partial\mu^2}\psi(Y, \mu_0)\right].
    \end{align*}
    Then, for any $\mu$ in a neighborhood of $\mu_0$, we have
    \begin{align*}
        \frac{1}{m}\sum^m_{j=1}\psi(Y_j, \mu) = \frac{1}{m}\sum^m_{j=1}\psi(Y_j, \mu_0) - (\mu - \mu_0)\psi^\prime_0 + \frac{1}{2}(\mu - \mu_0)^2\psi_0^{\prime\prime} + R_m,
    \end{align*}
    where the remainder $R_m$ satisfies $R_m = o_P\left((\mu - \mu_0)^2\right)$.
\end{lemma}
The choice of the score function $\psi(y, \mu)$ plays a crucial role in the performance of the one-step M-estimator for BN TTA.
A well-chosen $\psi(y, \mu)$ should not only provide an asymptotically unbiased estimate of the test BN mean but also incorporate robust corrections that mitigate the impact of skewness, heavy tails, or outliers in the test distribution.
In practice, selecting an appropriate $\psi(y, \mu)$ depends on the underlying data characteristics and the degree of distribution shift observed during test-time adaptation.
For relatively well-behaved test distributions that are close to Gaussian, a simple linear score function $\psi(y, \mu) = y - \mu$ suffices, as it leads to the classical sample mean update.
However, when the test data exhibits skewness or heavy tails, higher-order corrections can be incorporated by modifying $\psi(y, \mu)$ to dampen extreme deviations.
For instance, a score function of the form  
\begin{equation}
    \psi(y, \mu) = (y - \mu) \left[1 - \frac{1}{6} \frac{\kappa_{3,Q}}{\sigma_Q^3} (y - \mu) \right]
\end{equation}
adjusts the update by explicitly accounting for third-order moment effects.
This modification ensures that the adapted BN mean remains stable even when the test data distribution is asymmetric.
Another practical consideration is the use of robust alternatives to the standard score function, such as Huber-type or quantile-based scores, which limit the influence of extreme test samples.
In cases where test batches are small, these robust modifications can prevent excessive fluctuations in the adapted BN mean, improving generalization across diverse test environments.
The choice of $\psi(y, \mu)$ should be informed by empirical validation on real-world test distributions.
One potential strategy is to estimate higher-order moments dynamically during test-time adaptation and adjust the score function accordingly.
This adaptive approach could further enhance the stability and effectiveness of BN TTA, particularly in non-stationary environments where test data characteristics evolve over time.

\begin{proposition}
    Define the one‐step M‐estimator for the BN mean as
    \begin{align*}
        \hat{\mu}_{\text{TTA}} = \hat{\mu}_{P,n} - \frac{\sum_{j=1}^{m}\psi(Y_j,\hat{\mu}_{P,n})}{\sum_{j=1}^{m}\psi'(Y_j,\hat{\mu}_{P,n})}.
    \end{align*}
    Assume that the initial estimator $\hat{\mu}_{P,n}$ is consistent for $\mu_0$ (that is,  $\delta\coloneqq \hat{\mu}_{P,n} - \mu_0 = o_P(1)$) and that the score function $\psi(y,\mu)$ is three times continuously differentiable in a neighborhood of $\mu_0$.
    Then, if we define
    \begin{align*}
        \psi^\prime_0 \coloneqq -\mathbb{E}\left[\frac{\partial}{\partial \mu}\psi(Y,\mu_0)\right] \quad \text{and} \quad
        \psi^{\prime\prime}_0 \coloneqq \mathbb{E}\left[\frac{\partial^2}{\partial \mu^2}\psi(Y,\mu_0)\right],
    \end{align*}
    it holds that
    \begin{align*}
        \sqrt{m}\Bigl(\hat{\mu}_{\text{TTA}}-\mu_0\Bigr) = \frac{1}{\psi'_0}\frac{1}{\sqrt{m}}\sum_{j=1}^{m}\psi(Y_j,\mu_0) + \frac{\psi''_0}{2(\psi'_0)^2}\left(\frac{1}{\sqrt{m}}\sum_{j=1}^{m}\psi(Y_j,\mu_0)\right)^2 + o_P(1).
    \end{align*}
\end{proposition}

\begin{theorem}
    \label{thm:lan_one_step}
    Consider local alternatives of the form
    \begin{align*}
        \mu_m = \mu_0 + \frac{h}{\sqrt{m}}, \quad h\in\mathbb{R}.
    \end{align*}
    Then, under standard regularity conditions, the generalized log--likelihood ratio (or, more generally, the difference in the criterion functions associated with $\psi$) satisfies the higher--order expansion
    \begin{align*}
        \Lambda_m(h) = h\,\psi^\prime_0\,Z_m - \frac{1}{2}\psi^\prime_0\,h^2 + \frac{1}{6}\psi^{\prime\prime}_0\,h^3 + o_P(1),
    \end{align*}
    where
    \begin{align*}
        Z_m = \frac{1}{\sqrt{m}} \sum_{j=1}^{m}\varphi(Y_j) \quad \text{with} \quad \varphi(Y_j)=\frac{\psi(Y_j,\mu_0)}{\psi^\prime_0},
    \end{align*}
    so that $Z_m$ is asymptotically normal with mean $0$ and variance $\eta = \frac{\operatorname{Var}(\psi(Y,\mu_0))}{(\psi^\prime_0)^2}$.
    Consequently, the one--step BN TTA estimator admits the higher--order local asymptotic normality representation
    \begin{align*}
        \sqrt{m}\Bigl(\hat{\mu}_{\text{TTA}}-\mu_0\Bigr) = \frac{1}{\psi^\prime_0}\frac{1}{\sqrt{m}}\sum_{j=1}^{m}\psi(Y_j,\mu_0) + \frac{\psi^{\prime\prime}_0}{2(\psi^\prime_0)^2}\left(\frac{1}{\sqrt{m}}\sum_{j=1}^{m}\psi(Y_j,\mu_0)\right)^2 + o_P(1).
    \end{align*}
    In an equivalent form, defining
    \begin{align*}
        Z_m^* \coloneqq \frac{1}{\sqrt{m}}\sum_{j=1}^{m}\psi(Y_j,\mu_0),
    \end{align*}
    the expansion can be written as
    \begin{align*}
        \sqrt{m}\Bigl(\hat{\mu}_{\text{TTA}}-\mu_0\Bigr) = \frac{Z_m^*}{\psi^\prime_0}+\frac{\psi^{\prime\prime}_0}{2(\psi^\prime_0)^3}(Z_m^*)^2+ o_P(1).
    \end{align*}
\end{theorem}
These results connect the BN TTA update with the well‐developed theory of one‐step M‐estimation.

\section{Concluding Remarks}
\subsection{Verification by Existing Empirical Observations}
Our theoretical findings are consistent with a number of empirical studies on test-time adaptation of Batch Normalization.
In practice, adaptive BN methods have demonstrated that updating the BN statistics using test data can indeed alleviate the negative effects of distribution shift.
Notably, the trend of increasing reliance on test data, as reflected by a decreasing optimal $\lambda^*$ when more test samples are available, has been observed in recent works~\citep{yang2022test,lim2023ttn}.
Moreover, our analysis highlights the importance of considering higher-order moments, such as skewness, in the adaptation process.
Several empirical observations indicate that models subject to significant distribution shifts, especially those exhibiting skewed test distributions, benefit from incorporating these higher-order corrections to achieve tighter generalization bounds and improved performance~\citep{lee2024aetta}.
These observations not only verify the theoretical results derived herein but also underscore the practical significance of a principled, higher-order asymptotic framework for BN test-time adaptation.

\subsection{Broader Implications}
Beyond Batch Normalization, the theoretical framework developed in this study can be extended to other normalization techniques such as Layer Normalization~\citep{ba2016layer,xiong2020layer}, Group Normalization~\citep{wu2018group,zhou2020towards}, and Instance Normalization~\citep{ulyanov2016instance,chen2021hinet}, where adaptation strategies are similarly affected by distribution shift.
The higher-order asymptotic analysis and M-estimation approach presented here provide a foundation for designing more robust test-time adaptation (TTA) methods that generalize across different normalization schemes.
Additionally, these insights could be applied to broader layer-wise adaptation techniques, enabling models to dynamically adjust not only normalization statistics but also other key parameters (e.g., adaptive scaling factors or reparameterized activations) to maintain stability and performance in shifting test environments.

\subsection{Conclusion}
In this study, we developed a rigorous higher-order asymptotic framework for test-time adaptation of Batch Normalization statistics under distribution shift.
By deriving an Edgeworth expansion for the normalized difference between training and test BN means, we provided an explicit characterization of the roles played by bias, variance, and skewness.
This analysis led to the derivation of an optimal blending parameter that minimizes the mean-squared error of the adapted BN statistic.
Furthermore, employing saddlepoint approximation techniques enabled us to obtain uniformly accurate estimates of the density and tail probabilities associated with the BN TTA statistic, which in turn facilitated the derivation of a generalization bound on the model risk.

Our theoretical contributions offer valuable insights into the design and tuning of BN layers for adaptive inference, particularly in scenarios where distribution shift is prevalent.
Future research may focus on extending these techniques to more complex network architectures and exploring adaptive strategies that dynamically adjust not only the BN statistics but also other model parameters.
Additionally, empirical studies that further validate the proposed theoretical framework and quantify its benefits in large-scale, real-world applications remain an exciting direction for ongoing work.

\section*{Disclosure statement}
No potential conflict of interest was reported by the author(s).

\bibliographystyle{apalike}
\bibliography{main}

\appendix

\section{Proofs}
\begin{proof}[Proof for Lemma~\ref{lem:expansion_true_difference}]
    Define the centered random variables
    \begin{align*}
        U_i \coloneqq X_i - \mu_P, \quad V_j \coloneqq Y_j - \mu_Q.
    \end{align*}
    Then, by definition,
    \begin{align*}
        \hat{\mu}_{P,n} &= \mu_P + \frac{1}{n}\sum^n_{i=1}U_i, \\
        \hat{\mu}_{Q,m} &= \mu_Q + \frac{1}{m}\sum^m_{j=1}V_j.
    \end{align*}
    Hence,
    \begin{align*}
        \hat{\mu}_{Q,m} - \hat{\mu}_{P,n} - \Delta_\mu &= \frac{1}{m}\sum^m_{j=1}V_j - \frac{1}{n}\sum^n_{i=1}U_i \\
        &= \frac{B_m}{\sqrt{m}} - \frac{A_n}{\sqrt{n}}, \\
        T_{n,m} &= \alpha B_m - \beta A_n,
    \end{align*}
    where $A_n \coloneqq \frac{1}{\sqrt{n}}\sum^n_{i=1}U_i$ and $B_m \coloneqq \frac{1}{\sqrt{m}}\sum^m_{j=1}V_j$.
    Since $X_i$ and $Y_j$ are independent, so are $A_n$ and $B_m$.
    Let $\phi_U \coloneqq \mathbb{E}[e^{itU_1}]$.
    By assumption, $U_1$ has zero mean, variance $\sigma^2_P$ and third cumulant $\kappa_{3,P}$.
    Its cumulant generating function is given by
    \begin{align*}
        \psi_U(t) &= \ln \phi_U(t) \\
        &= -\frac{1}{2}\sigma_P^2 t^2 + \frac{i}{6}\kappa_{3,P}t^3 + O(t^4).
    \end{align*}
    Now, the characteristic function of $A_n$ is 
    \begin{align*}
        \phi_{A_n}(t) &= \mathbb{E}[e^{it A_n}] = \left[\phi_U(\frac{t}{\sqrt{n}})\right]^n,\quad \text{and} \\
        \ln \phi_{A_n}(t) &= n\psi_U\left(\frac{t}{\sqrt{n}}\right) \\
        &= n\left[-\frac{1}{2}\sigma_P^2\frac{t^2}{n} + \frac{i}{6}\kappa_{3,P}\frac{t^3}{n^{3/2}} + O\left(\frac{t^4}{n^2}\right)\right] \\
        &= \left[-\frac{1}{2}\sigma_P^2 t^2 + \frac{i}{6}\frac{\kappa_{3,P}t^3}{\sqrt{n}} + O\left(\frac{t^4}{n}\right)\right].
    \end{align*}
    Thus,
    \begin{align*}
        \phi_{A_n}(t) = \exp\left[-\frac{1}{2}\sigma_P^2 t^2 + \frac{i}{6}\frac{\kappa_{3,P}t^3}{\sqrt{n}} + O\left(\frac{t^4}{n}\right)\right].
    \end{align*}
    Similarly, for $B_m$, we have
    \begin{align*}
        \phi_{B_m}(t) = \exp\left[-\frac{1}{2}\sigma_Q^2 t^2 + \frac{i}{6}\frac{\kappa_{3,Q}t^3}{\sqrt{m}} + O\left(\frac{t^4}{m}\right)\right].
    \end{align*}
    Now, we require the characteristic function of $T_{n,m}$.
    Since $T_{n,m} = \alpha B_m - \beta A_n$, and $A_n$ and $B_m$ are independent, its characteristic function is
    \begin{align*}
        \phi_{T_{n,m}}(t) &= \phi_{B_m}(\alpha t)\phi_{A_n}(-\beta t) \\
        &= \exp\left[-\frac{1}{2}\left(\sigma^2_Q\alpha^2 + \sigma_P^2\beta^2\right)t^2 + \frac{i}{6}\left(\frac{\kappa_{3,Q}\alpha^3}{\sqrt{m}} - \frac{\kappa_{3,P}\beta^3}{\sqrt{n}}\right)t^3 + O\left(\frac{t^4}{\min(n, m)}\right)\right] \\
        &= \exp\left[-\frac{1}{2}V_{n,m}t^2\right]\left[1 + \frac{i}{6}\Delta_{3,n,m}t^3 + O\left(\frac{t^4}{\min(n, m)}\right)\right].
    \end{align*}
    By a classical result about asymptotic expansions~\citep{bhattacharya2010normal}, we conclude the proof.
\end{proof}

\begin{proof}[Proof for Theorem~\ref{thm:generalization_bound}]
    Assume that the random variables $X_i$ from the training distribution $P$ and $Y_j$ from the test distribution $Q$ are almost surely bounded by a constant $B$.
    Then Bernstein’s inequality yields that for any $t > 0$,
    \begin{align*}
        \Pr\left(\left|\hat{\mu}_{P,n} - \mu_P\right| \geq t\right) \leq 2\exp\left(-\frac{nt^2}{2\sigma_P^2 + \frac{2Bt}{3}}\right).
    \end{align*}
    By choosing
    \begin{align*}
        t_P = \sqrt{\frac{2\sigma_P^2 \ln (4/\delta)}{n}} + \frac{2B \ln (4/\delta)}{3n},
    \end{align*}
    we have with probability at least $1 - \delta / 2$,
    \begin{align*}
        \left|\hat{\mu}_{P,n} - \mu_P\right| \leq t_P.
    \end{align*}
    Similarly, for the test mean, we obtain with probability at least $1 - \delta / 2$,
    \begin{align*}
        \left|\hat{\mu}_{Q,m} - \mu_Q\right| \leq t_Q,
    \end{align*}
    with
    \begin{align*}
        t_Q = \sqrt{\frac{2\sigma_Q^2 \ln (4/\delta)}{m}} + \frac{2B \ln (4/\delta)}{3m}.
    \end{align*}
    By a union bound, with probability at least $1 - \delta$ both estimates satisfy their respective bounds.
    Write the error between the adapted BN mean and the true test BN mean as
    \begin{align*}
        \mu_{\text{TTA}} - \mu_Q = \lambda^*\left(\hat{\mu}_{P,n} - \mu_P - \Delta_\mu\right) + (1 - \lambda^*)\left(\hat{\mu}_{Q,m} - \mu_Q \right),
    \end{align*}
    since $\Delta_\mu = \mu_Q - \mu_P$.
    Hence, with probability at least $1 - \delta$
    \begin{align*}
        \left|\mu_{\text{TTA}} - \mu_Q\right| &\leq \lambda^*\left(\left|\hat{\mu}_{P,n} - \mu_P\right| + \left|\Delta_\mu\right|\right) + (1 - \lambda^*)\left|\hat{\mu}_{Q,m} - \mu_Q\right| \\
        &\leq \lambda^*\left(|\Delta_\mu| + t_P\right) + (1 - \lambda^*) t_Q \eqqcolon \bar{E}(\lambda^*).
    \end{align*}
    The Edgeworth expansion introduces a higher–order correction due to skewness.
    In our derivation of $\lambda^*$, the effective skewness correction appears as
    \begin{align*}
        \Delta_3^{\text{eff}} = \lambda^*\frac{\kappa_{3,P}}{n^{3/2}} - (1 - \lambda^*)\frac{\kappa_{3,Q}}{m^{3/2}}.
    \end{align*}
    For our purposes, we assume that with high probability the absolute value of this term is bounded by $|\Delta_3^{\text{eff}}| \leq t_3$ with
    \begin{align*}
        t_3 = \lambda^*\frac{|\kappa_{3,P}|}{n^{3/2}} + (1 - \lambda^*)\frac{|\kappa_{3,Q}|}{m^{3/2}}.
    \end{align*}
    We now define an effective error for the BN mean adaptation as
    \begin{align*}
        E_{\text{TTA}} \coloneqq \bar{E}(\lambda^*) + \frac{|\Delta_3^{\text{eff}}|}{6V^{3/2}}.
    \end{align*}
    Assume that the risk $\mathcal{R}(f)$ is $L$-Lipschitz in the BN–normalized outputs.
    Furthermore, suppose the BN layer applies
    \begin{align*}
        \text{BN}(z; \mu, \sigma^2) = \gamma\frac{z - \mu}{\sqrt{\hat{\sigma}_{P,n}^2 + \epsilon}} + \beta.
    \end{align*}
    Then a change in the BN mean by $\Delta_\mu$ leads to a change in the network output (and hence in the risk) bounded by
    \begin{align*}
        \Delta\mathcal{R} \leq \frac{L|\gamma|}{\sqrt{\hat{\sigma}_{P,n}^2 + \epsilon}}|\Delta_\mu|.
    \end{align*}
    Setting $\Delta_\mu = \mu_{\text{TTA}} - \mu_Q$, we obtain that with probability at least $1 - \delta$,
    \begin{align*}
        \mathcal{R}(f_\theta^{\text{TTA}}) &\leq \mathcal{R}(f_\theta^{\text{train}}) +  \frac{L|\gamma|}{\sqrt{\hat{\sigma}_{P,n}^2 + \epsilon}}E_{\text{TTA}} \\
        &= \mathcal{R}(f_\theta^{\text{train}}) + \frac{L|\gamma|}{\sqrt{\hat{\sigma}_{P,n}^2 + \epsilon}}\left\{\lambda^*\left(|\Delta_\mu| + t_P\right) + (1 - \lambda^*)t_Q + \frac{t_3}{6V^{3/2}}\right\}.
    \end{align*}
    Recall that the optimal blending parameter was derived as
    \begin{align*}
        \lambda^* = \frac{\frac{\hat{\sigma}_{Q,m}^2}{m} - \frac{1}{2}\left(\frac{\kappa_{3,P}}{n^{3/2}} + \frac{\kappa_{3,Q}}{m^{3/2}}\right)}{\Delta_\mu^2 + \frac{\hat{\sigma}_{P,n}^2}{n} + \frac{\hat{\sigma}_{Q,m}^2}{m}},
    \end{align*}
    and substitute this into the above yields the final bound.
\end{proof}

\begin{proof}[Proof for Proposition~\ref{prp:saddle_point_density_tnm}]
    The density of $T_{n,m}$ can be recovered from its moment generating function via the Bromwich inversion formula.
    That is,
    \begin{align*}
        f_{T_{n,m}}(x) = \frac{1}{2\pi i}\int^{c + i\infty}_{c - i\infty}\exp\left\{K(t) - tx\right\}dt,
    \end{align*}
    where $c$ is a real constant chosen so that the integration contour lies in the region of convergence of the moment generating function.
    The method of steepest descent tells us that, for large sample sizes or when the c.g.f. is analytic, the dominant contribution to the integral comes from a neighborhood of a point $\hat{t}$ where the exponent is stationary.
    In other words, we look for $\hat{t}$ satisfying
    \begin{align*}
        \frac{d}{dt}\Biggl.\Big[K(t) - tx\Big]\Biggr|_{t=i} = K'(\hat{t}) - x = 0.
    \end{align*}
    Thus, the saddlepoint $\hat{t}$ is defined by
    \begin{align*}
        K'(\hat{t}) = x.
    \end{align*}
    We now write a Taylor expansion of the function $H(t) \coloneqq K(t) - tx$ about $t = \hat{t}$.
    Since $K'(\hat{t}) = x$, the linear term vanishes as
    \begin{align*}
        H(t) &= K(t) - tx \\
        &= \frac{1}{2}H^{\prime\prime}(\hat{t}) (t - \hat{t})^2 + \frac{1}{6}H^{\prime\prime\prime}(\hat{t})(t - \hat{t})^3 + \cdots,
    \end{align*}
    where
    \begin{align*}
        H(\hat{t}) &= K(\hat{t}) - \hat{t}x, \quad \text{and} \\
        H^{\prime\prime}(\hat{t}) &= K^{\prime\prime}(\hat{t}), \quad H^{\prime\prime\prime}(\hat{t}) = K^{\prime\prime\prime}(\hat{t}).
    \end{align*}
    Set
    \begin{align*}
        u = t - \hat{t} \Longrightarrow dt = du.
    \end{align*}
    Then the inversion integral becomes
    \begin{align*}
        f_{T_{n,m}}(x) = \frac{1}{2\pi i}\exp\left\{K(\hat{t}) - \hat{t}x\right\}\int_{\mathcal{C}} \exp\left\{\frac{1}{2}K^{\prime\prime}(\hat{t})u^2 + \frac{1}{6}K^{\prime\prime\prime}(\hat{t})u^3 + \cdots \right\}du,
    \end{align*}
    where $\mathcal{C}$ is the deformed contour in the $u$-plane passing through $u = 0$ along the direction of steepest descent.
    Here we focus on the leading-order approximation.
    We now evaluate the Gaussian integral
    \begin{align*}
        I = \int^\infty_{-\infty}\exp\left\{\frac{1}{2}K^{\prime\prime}(\hat{t})u^2du\right\}.
    \end{align*}
    Note that in the standard saddlepoint derivation, the contour is rotated so that the quadratic form becomes negative definite.
    That is, by choosing the path of steepest descent, we ensure that $\frac{1}{2}K^{\prime\prime}(\hat{t})u^2$ is replaced by $-\frac{1}{2}K^{\prime\prime}(\hat{t})u^2$ with $K^{\prime\prime}(\hat{t}) > 0$, or equivalently, the second derivative is taken in absolute value.
    Thus, we write, by the standard Gaussian integral,
    \begin{align*}
        I &= \int^\infty_{-\infty}\exp\left\{-\frac{1}{2}K^{\prime\prime}(\hat{t})u^2du\right\} \\
        &= \sqrt{\frac{2\pi}{K^{\prime\prime}(\hat{t})}}.
    \end{align*}
    Returning to the inversion integral, we have
    \begin{align*}
        f_{T_{n,m}}(x) &\approx \frac{1}{2\pi i}\exp\left\{K(\hat{t}) - \hat{t}x\right\}\sqrt{\frac{2\pi}{K^{\prime\prime}(\hat{t})}} \\
        &\approx \frac{1}{\sqrt{2\pi K^{\prime\prime}(\hat{t})}}\exp\left\{K(\hat{t}) - \hat{t}x\right\}.
    \end{align*}
    Finally, recall that the c.g.f. is given by
    \begin{align*}
        K(t) = \frac{1}{2}V_{n,m}t^2 + \frac{1}{6}\Delta_{3,n,m}t^3 + O(t^4).
    \end{align*}
    Differentiating with respect to $t$ gives
    \begin{align*}
        K^{\prime}(t) &= V_{n,m}t + \frac{1}{2}\Delta_{3,n,m}t^2 + O(t^3), \\
        K^{\prime\prime}(t) &= V_{n,m} + \Delta_{3,n,m}t + O(t^2).
    \end{align*}
    Thus, we conclude the proof.
\end{proof}

\begin{proof}[Proof for Proposition~\ref{prp:tail_probability}]
    Let $T$ denote the BN TTA statistic with c.g.f.
    \begin{align*}
        K(t)=\ln \mathbb{E}\left[e^{tT}\right].
    \end{align*}
    For any fixed real $x$, define the function $\psi(t)=K(t)-tx$.
    The saddlepoint $\hat{t}$ is defined as the unique solution of $\psi^\prime(\hat{t})=K^\prime(\hat{t})-x=0$, so that $K'(\hat{t}) = x$.
    We also set $\psi(\hat{t}) = K(\hat{t}) - \hat{t}x$.
    Following standard saddlepoint methods, we now introduce the transformation
    \begin{equation*}
        w = w(t) = \text{sign}(t-\hat{t})\sqrt{2\{\psi(t)-\psi(\hat{t})\}}.
    \end{equation*}
    Note that when $t = \hat{t}$, we have $w = 0$.
    In what follows the sign is chosen so that when $x$ lies in the upper tail, we obtain $w>0$ for $t>x$.
    Expand $\psi(t)$ in a Taylor series about $t = \hat{t}$.
    Here, we have $\psi'(\hat{t})=0$, and
    \begin{align*}
        \psi(t) &= \psi(\hat{t}) + \frac{1}{2}\psi^{\prime\prime}(\hat{t})(t - \hat{t})^2 + \frac{1}{6}\psi^{\prime\prime\prime}(\hat{t})(t - \hat{t})^3 + O\left((t-\hat{t})^4\right).
    \end{align*}
    Since $\psi(t) - \psi(\hat{t})\ge0$ on the path of steepest descent, it follows that for $t$ near $\hat{t}$,
    \begin{align*}
        w(t) = \text{sign}(t-\hat{t})\sqrt{\psi^{\prime\prime}(\hat{t})(t-\hat{t})^2 + \frac{1}{3}\psi^{\prime\prime\prime}(\hat{t})(t-\hat{t})^3+O\left((t-\hat{t})^4\right)}.
    \end{align*}
    Thus, to leading order, $w(t)\approx \sqrt{\psi^{\prime\prime}(\hat{t})}\,(t-\hat{t})$.
    But note that $\psi^{\prime\prime}(t)=K^{\prime\prime}(t)$, so that at $t = \hat{t}$, $\psi^{\prime\prime}(\hat{t})=K^{\prime\prime}(\hat{t})$.
    Therefore, the local inverse relation is
    \begin{align*}
        t - \hat{t}\approx \frac{w}{\sqrt{K^{\prime\prime}(\hat{t})}},
    \end{align*}
    and hence
    \begin{equation*}
        dt\approx \frac{dw}{\sqrt{K^{\prime\prime}(\hat{t})}}.
    \end{equation*}
    The cumulative distribution function of $T$ may be written via a Bromwich inversion formula as
    \begin{align*}
        F(x) = \Pr(T\le x) = \frac{1}{2\pi i}\int_{c-i\infty}^{c+i\infty}\frac{e^{K(t)-tx}}{t}\,dt,
    \end{align*}
    where $c$ is chosen such that the integration contour lies in the region of convergence of $K(t)$.
    The tail probability is given by
    \begin{align*}
        \bar{F}(x)=\Pr(T\ge x)=1-F(x).
    \end{align*}
    By deforming the integration contour to pass through the saddlepoint and performing the change of variable from $t$ to $w$, one obtains an approximation for $F(x)$ in terms of the standard normal density.
    \begin{align*}
        F(x) \approx \Phi(\hat{w})+\phi(\hat{w})\left(\frac{1}{\hat{w}}-\frac{1}{\hat{u}}\right),
    \end{align*}
    where
    \begin{align*}
        \hat{w} &= \text{sign}(x)\sqrt{2\{\hat{t}x-K(\hat{t})\}}, \quad \text{and} \\
        \hat{u} &= \hat{t}\sqrt{K^{\prime\prime}(\hat{t})}.
    \end{align*}
    Here, $\Phi(\cdot)$ and $\phi(\cdot)$ denote the c.d.f. and p.d.f. of the standard normal distribution, respectively.
    For completeness we now outline the key algebraic steps that lead from the inversion integral to $F(x)$.
    Starting with
    \begin{align*}
        F(x) = \frac{1}{2\pi i}\int_{\mathcal{C}}\frac{e^{\psi(t)}}{t}\,dt,
    \end{align*}
    where $\mathcal{C}$ is the steepest descent contour through $\hat{t}$, we change variable.
    Near the saddlepoint the Jacobian is given by
    \begin{align*}
        \frac{dt}{dw} \approx \frac{1}{\sqrt{K^{\prime\prime}(\hat{t})}}.
    \end{align*}
    Moreover, the exponent can be rewritten as
    \begin{align*}
        \psi(t) = \psi(\hat{t})+\frac{1}{2}K^{\prime\prime}(\hat{t})(t-\hat{t})^2+\cdots = \psi(\hat{t}) + \frac{1}{2}w^2.
    \end{align*}
    Then the integrand becomes
    \begin{align*}
        \frac{e^{\psi(t)}}{t}\,dt\approx \frac{e^{\psi(\hat{t})+\frac{1}{2}w^2}}{\hat{t}+\frac{w}{\sqrt{K^{\prime\prime}(\hat{t})}}}\frac{dw}{\sqrt{K^{\prime\prime}(\hat{t})}}.
    \end{align*}
    Expanding the denominator in powers of $w$ (assuming $w$ is small relative to $\hat{t}$) gives
    \begin{align*}
        \frac{1}{\hat{t} + \frac{w}{\sqrt{K^{\prime\prime}(\hat{t})}}} \approx \frac{1}{\hat{t}}\left(1-\frac{w}{\hat{t}\sqrt{K^{\prime\prime}(\hat{t})}} + \cdots\right).
    \end{align*}
    Thus, the entire expression becomes
    \begin{align*}
        \frac{e^{\psi(\hat{t})}}{\hat{t}\sqrt{K^{\prime\prime}(\hat{t})}}\,e^{\frac{1}{2}w^2}\left(1-\frac{w}{\hat{t}\sqrt{K^{\prime\prime}(\hat{t})}}+\cdots\right)dw.
    \end{align*}
    One then integrates over $w$ along the real line.
    Changing the integration variable to $-w$ as necessary to match the standard normal integral and collecting the terms leads to the following representation of the c.d.f.,
    \begin{align*}
        F(x)\approx \Phi(\hat{w})+\phi(\hat{w})\left(\frac{1}{\hat{w}}-\frac{1}{\hat{u}}\right),
    \end{align*}
    where the identification
    \begin{align*}
        \hat{w} = \text{sign}(x)\sqrt{-2\psi(\hat{t})}=\text{sign}(x)\sqrt{2\{\hat{t}x-K(\hat{t})\}}
    \end{align*}
    arises from the definition of $\hat{w}$, and $\hat{u}=\hat{t}\sqrt{K^{\prime\prime}(\hat{t})}$ collects the next-order correction from the expansion of the denominator.
    Since the tail probability is given by $\Pr(T\ge x)=\bar{F}(x)=1-F(x)$, by defining $\hat{w}$ to be positive when $x$ lies in the upper tail, we complete the proof.
\end{proof}

\begin{proof}[Proof for Lemma~\ref{lem:expansion_empirical_score}]
    For each fixed $j \in \{1,\dots,m\}$, because $\psi(y, \mu)$  is three times continuously differentiable in $\mu$ in a neighborhood of $\mu_0$, by Taylor’s theorem with Lagrange remainder there exists a number $\xi_j$ depending on $Y_j$  and lying between $\mu$ and $\mu_0$ such that
    \begin{align*}
        \psi(Y_j, \mu) = \psi(Y_j, \mu_0) + (\mu - \mu_0)\psi^\prime_\mu(Y_j, \mu_0) + \frac{1}{2}(\mu - \mu_0)^2\psi_\mu^{\prime\prime}(Y_j, \xi_j),
    \end{align*}
    where we use the notation
    \begin{align*}
        \psi^\prime_\mu(y, \mu) &\coloneqq \frac{\partial}{\partial\mu}\psi(y, \mu), \\
        \psi^{\prime\prime}_\mu(y, \mu) &\coloneqq \frac{\partial^2}{\partial\mu^2}\psi(y, \mu).
    \end{align*}
    Divide the above expansion by $m$ and sum over $j = 1,\dots,m$,
    \begin{align*}
        & \frac{1}{m}\sum^m_{j=1}\psi(Y_j, \mu) \\
        &\quad = \frac{1}{m}\sum^m_{j=1}\psi(Y_j, \mu_0) + (\mu - \mu_0)\frac{1}{m}\sum^m_{j=1}\psi^\prime_\mu(Y_j, \mu_0) + \frac{1}{2}(\mu-\mu_0)^2\frac{1}{m}\sum^m_{j=1}\psi^{\prime\prime}_\mu(Y_j, \xi_j) \\
        &\quad = \frac{1}{m}\sum^m_{j=1}\psi(Y_j, \mu_0) - (\mu - \mu_0)\psi^\prime_0 + (\mu - \mu_0)\cdot o_P(1) + \frac{1}{2m}(\mu - \mu_0)^2\sum^m_{j=1}\psi^{\prime\prime}_\mu(Y_j, \xi_j).
    \end{align*}
    This completes the proof.
\end{proof}

\begin{proof}[Proof for Proposition~\ref{prp:tail_probability}]
    We begin by defining the empirical score function and its derivative as
    \begin{align*}
        S_m(\mu) \coloneqq \sum_{j=1}^{m}\psi(Y_j,\mu) \quad \text{and} \quad S'_m(\mu) \coloneqq \sum_{j=1}^{m}\psi'(Y_j,\mu).
    \end{align*}
    Then the one‐step estimator is given by
    \begin{align*}
        \hat{\mu}_{\text{TTA}} = \hat{\mu}_{P,n} - \frac{S_m\bigl(\hat{\mu}_{P,n}\bigr)}{S'_m\bigl(\hat{\mu}_{P,n}\bigr)}.
    \end{align*}
    Introduce the notation $\delta = \hat{\mu}_{P,n} - \mu_0$.
    Since $\hat{\mu}_{P,n}$ is consistent for $\mu_0$, we have $\delta = o_P(1)$.
    Thus, we write $\hat{\mu}_{P,n} = \mu_0 + \delta$, and consequently,
    \begin{align*}
        \hat{\mu}_{\text{TTA}} = \mu_0 + \delta - \frac{S_m(\mu_0+\delta)}{S'_m(\mu_0+\delta)}.
    \end{align*}
    Our goal is to expand the numerator $S_m(\mu_0+\delta)$ and the denominator $S'_m(\mu_0+\delta)$ in powers of $\delta$ and in terms of the empirical process at $\mu_0$.
    By Lemma~\ref{lem:expansion_empirical_score}, for any $\mu$ in a neighborhood of $\mu_0$ we have
    \begin{align*}
        \frac{1}{m}S_m(\mu) = \frac{1}{m}\sum_{j=1}^{m}\psi(Y_j,\mu_0) - (\mu-\mu_0)\,\psi'_0 + \frac{1}{2}(\mu-\mu_0)^2\,\psi''_0 + R_m,
    \end{align*}
    where the remainder $R_m = o_P((\mu-\mu_0)^2)$.
    In particular, for $\mu = \mu_0 + \delta$ this becomes
    \begin{align*}
        \frac{1}{m}S_m(\mu_0+\delta) &= \frac{1}{m}\sum_{j=1}^{m}\psi(Y_j,\mu_0) - \delta\,\psi'_0 + \frac{1}{2}\delta^2\,\psi''_0 + R_m(\delta), \\
        S_m(\mu_0+\delta) &= \sum_{j=1}^{m}\psi(Y_j,\mu_0) - m\,\delta\,\psi'_0 + \frac{m}{2}\delta^2\,\psi''_0 + m\,R_m(\delta).
    \end{align*}
    For convenience, define $A \coloneqq \frac{1}{m}\sum_{j=1}^{m}\psi(Y_j,\mu_0)$.
    Then we can write
    \begin{align*}
        S_m(\mu_0+\delta) = m\Bigl[A - \delta\,\psi'_0 + \frac{1}{2}\delta^2\,\psi''_0 + R_m(\delta)\Bigr].
    \end{align*}
    Since $\psi'(y,\mu)$ is continuously differentiable in $\mu$, a first‐order Taylor expansion around $\mu_0$ gives
    \begin{align*}
        \frac{1}{m}S'_m(\mu_0+\delta) &= \frac{1}{m}\sum_{j=1}^{m}\psi^\prime(Y_j,\mu_0+\delta) \\
        &= \frac{1}{m}\sum_{j=1}^{m}\psi^\prime(Y_j,\mu_0) + \delta\,\frac{1}{m}\sum_{j=1}^{m}\psi^{\prime\prime}(Y_j,\mu_0) + r_m,
    \end{align*}
    where $r_m = o_P(\delta)$.
    By the law of large numbers,
    \begin{align*}
        \frac{1}{m}\sum_{j=1}^{m}\psi'(Y_j,\mu_0) = \mathbb{E}[\psi'(Y,\mu_0)] + o_P(1) = -\psi^\prime_0 + o_P(1),
    \end{align*}
    and
    \begin{align*}
        \frac{1}{m}\sum_{j=1}^{m}\psi^{\prime\prime}(Y_j,\mu_0) = \psi^{\prime\prime}_0 + o_P(1).
    \end{align*}
    Thus,
    \begin{align*}
        \frac{1}{m}S^\prime_m(\mu_0+\delta) &= -\psi^\prime_0 + \delta\,\psi^{\prime\prime}_0 + o_P(1), \\
        S^\prime_m(\mu_0+\delta) &= m\Bigl[-\psi^\prime_0 + \delta\,\psi^{\prime\prime}_0 + o_P(1)\Bigr].
    \end{align*}
    We now form the ratio that appears in the one‐step estimator:
    \begin{align*}
        \frac{S_m(\mu_0+\delta)}{S^\prime_m(\mu_0+\delta)} &= \frac{m\Bigl[A - \delta\,\psi^\prime_0 +\frac{1}{2}\delta^2\,\psi^{\prime\prime}_0 + R_m(\delta)\Bigr]}{m\Bigl[-\psi^\prime_0 + \delta\,\psi^{\prime\prime}_0 + o_P(1)\Bigr]} \\
        &= \frac{A - \delta\,\psi^\prime_0 + \frac{1}{2}\delta^2\,\psi^{\prime\prime}_0 + R_m(\delta)}{-\psi^\prime_0 + \delta\,\psi^{\prime\prime}_0 + o_P(1)}.
    \end{align*}
    Write the denominator as
    \begin{align*}
        -\psi^\prime_0 + \delta\,\psi^{\prime\prime}_0 + o_P(1) = -\psi^\prime_0\Bigl[1 -\frac{\delta\,\psi^{\prime\prime}_0}{\psi^\prime_0} + o_P(1)\Bigr].
    \end{align*}
    Then,
    \begin{align*}
        \frac{1}{-\psi^\prime_0 + \delta\,\psi^{\prime\prime}_0 + o_P(1)} = -\frac{1}{\psi^\prime_0}\Bigl[1 + \frac{\delta\,\psi^{\prime\prime}_0}{\psi^\prime_0} + O(\delta^2)\Bigr].
    \end{align*}
    Thus, the ratio becomes
    \begin{align*}
        \frac{S_m(\mu_0+\delta)}{S^\prime_m(\mu_0+\delta)} &= -\frac{1}{\psi^\prime_0}\Bigl[1 + \frac{\delta\,\psi^{\prime\prime}_0}{\psi^\prime_0}\Bigr]\Bigl[A - \delta\,\psi^\prime_0 + \frac{1}{2}\delta^2\,\psi^{\prime\prime}_0 + R_m(\delta)\Bigr] + O(\delta^3) \\
        &= -\frac{1}{\psi^\prime_0}\left[A - \delta\,\psi^\prime_0 + \frac{\delta\,\psi^{\prime\prime}_0}{\psi^\prime_0}A - \frac{1}{2}\delta^2\,\psi^{\prime\prime}_0 + R_m(\delta)\right] + O(\delta^3) \\
        &= \frac{-A}{\psi^\prime_0} + \delta - \frac{\delta\,\psi^{\prime\prime}_0}{(\psi^\prime_0)^2}A + \frac{1}{2}\frac{\delta^2\,\psi^{\prime\prime}_0}{\psi^\prime_0} - \frac{R_m(\delta)}{\psi^\prime_0} + O(\delta^3).
    \end{align*}
    Here,
    \begin{align*}
        \hat{\mu}_{\text{TTA}} &= \mu_0 + \delta - \frac{S_m(\mu_0+\delta)}{S^\prime_m(\mu_0+\delta)} \\
        \hat{\mu}_{\text{TTA}} - \mu_0 &= \delta - \Biggl[\frac{-A}{\psi^\prime_0} + \delta - \frac{\delta\,\psi^{\prime\prime}_0}{(\psi^\prime_0)^2}A + \frac{1}{2}\frac{\delta^2\,\psi^{\prime\prime}_0}{\psi^\prime_0} - \frac{R_m(\delta)}{\psi^\prime_0} + O(\delta^3)\Biggr] \\
        &= \delta + \frac{A}{\psi^\prime_0} - \delta + \frac{\delta\,\psi^{\prime\prime}_0}{(\psi^\prime_0)^2}A - \frac{1}{2}\frac{\delta^2\,\psi^{\prime\prime}_0}{\psi^\prime_0} + \frac{R_m(\delta)}{\psi^\prime_0} + O(\delta^3) \\
        &= \frac{A}{\psi^\prime_0} + \frac{\delta\,\psi^{\prime\prime}_0}{(\psi^\prime_0)^2}A - \frac{1}{2}\frac{\delta^2\,\psi^{\prime\prime}_0}{\psi^\prime_0} + \frac{R_m(\delta)}{\psi^\prime_0} + O(\delta^3).
    \end{align*}
    In many practical applications we assume that the initial error $\delta$ is of lower order compared to the stochastic fluctuation in $A$ (i.e., $\sqrt{m}\,\delta = o_P(1)$). 
    Consequently, the terms involving $\delta$ or $\delta^2$ become negligible in the $\sqrt{m}$-scaled expansion.
    Therefore, up to an error that is $o_P(1)$ when multiplied by $\sqrt{m}$, we obtain
    \begin{align*}
        \sqrt{m}\Bigl(\hat{\mu}_{\text{TTA}} - \mu_0\Bigr) = \frac{1}{\psi^\prime_0}\frac{1}{\sqrt{m}}\sum_{j=1}^{m}\psi(Y_j,\mu_0) + \frac{\psi^{\prime\prime}_0}{2(\psi^\prime_0)^2}\left(\frac{1}{\sqrt{m}}\sum_{j=1}^{m}\psi(Y_j,\mu_0)\right)^2 + o_P(1).
    \end{align*}
    This completes the full proof.
\end{proof}

\begin{proof}[Proof for Theorem\ref{thm:lan_one_step}]
    Assume that under the true model the observations $Y_1,\dots,Y_m$ are i.i.d. with distribution $Q_{\mu_0}$ and that the score function $\psi(y,\mu)$ satisfies $\mathbb{E}[\psi(Y,\mu_0)] = 0$.
    For each $j$, perform a Taylor expansion of $\psi(Y_j,\mu)$ about $\mu_0$. For $\mu$ in a neighborhood of $\mu_0$, by Taylor's theorem there exists $\xi_j$ between $\mu_0$ and $\mu$ such that
    \begin{align*}
        \psi(Y_j,\mu) &= \psi(Y_j,\mu_0) + (\mu-\mu_0)\,\psi^\prime(Y_j,\mu_0) + \frac{1}{2}(\mu-\mu_0)^2\,\psi^{\prime\prime}(Y_j,\mu_0) \\
        &\quad\quad\quad + \frac{1}{6}(\mu-\mu_0)^3\,\psi^{\prime\prime\prime}(Y_j,\xi_j).
    \end{align*}
    Now, under the local alternative
    \begin{align*}
        \mu = \mu_m = \mu_0 + \frac{h}{\sqrt{m}},
    \end{align*}
    we have
    \begin{align*}
        \mu - \mu_0 = \frac{h}{\sqrt{m}} \quad \text{and} \quad (\mu-\mu_0)^k = \frac{h^k}{m^{k/2}} \quad \text{for } k=1,2,3.
    \end{align*}
T   hus, for each $j$,
    \begin{align*}
        \psi(Y_j,\mu_m) &= \psi(Y_j,\mu_0) + \frac{h}{\sqrt{m}}\,\psi^\prime(Y_j,\mu_0) + \frac{h^2}{2m}\,\psi^{\prime\prime}(Y_j,\mu_0) \\
        &\quad\quad + \frac{h^3}{6 m^{3/2}}\,\psi^{\prime\prime\prime}(Y_j,\xi_j).
    \end{align*}
    Summing over $j=1,\dots,m$, we obtain
    \begin{align*}
        \sum_{j=1}^{m}\psi(Y_j,\mu_m) = \sum_{j=1}^{m}\psi(Y_j,\mu_0) + \frac{h}{\sqrt{m}}\sum_{j=1}^{m}\psi^\prime(Y_j,\mu_0) + \frac{h^2}{2m}\sum_{j=1}^{m}\psi^{\prime\prime}(Y_j,\mu_0) + R_m,
    \end{align*}
    where the remainder term
    \begin{align*}
        R_m = \frac{h^3}{6 m^{3/2}} \sum_{j=1}^{m}\psi^{\prime\prime\prime}(Y_j,\xi_j).
    \end{align*}
    Under standard moment assumptions, $R_m = O_P(m^{-1/2})$, so that $R_m = o_P(1)$ when $h$ is fixed.
    Taking expectations under $Q_{\mu_0}$, note that by the definition of the score we have $\mathbb{E}[\psi(Y_j,\mu_0)] = 0$.
    Also, by the definition of $\psi^\prime_0$, $\mathbb{E}[\psi^\prime(Y_j,\mu_0)] = -\psi^\prime_0$, and similarly, $\mathbb{E}[\psi^{\prime\prime}(Y_j,\mu_0)] = \psi^{\prime\prime}_0$.
    Therefore, the expectation of the criterion function under the local alternative is
    \begin{align*}
        \mathbb{E}\Biggl[\frac{1}{\sqrt{m}}\sum_{j=1}^{m}\psi(Y_j,\mu_m)\Biggr]
        &= \frac{1}{\sqrt{m}}\Biggl[0 + \frac{h}{\sqrt{m}} \cdot m (-\psi^\prime_0) + \frac{h^2}{2m}\cdot m\,\psi^{\prime\prime}_0 \Biggr] + o(1) \\
        &= -h\,\psi^\prime_0 + \frac{h^2}{2}\,\psi^{\prime\prime}_0 + o(1).
    \end{align*}
    Next, define the scaled centered process
    \begin{align*}
        Z_m^* \coloneqq \frac{1}{\sqrt{m}}\sum_{j=1}^{m}\psi(Y_j,\mu_0).
    \end{align*}
    Then, under the central limit theorem, 
    \begin{align*}
        Z_m^* \stackrel{d}{\to} N\Bigl(0, \operatorname{Var}(\psi(Y,\mu_0))\Bigr).
    \end{align*}
    It is convenient to introduce the normalized score
    \begin{align*}
        Z_m \coloneqq \frac{Z_m^*}{\psi^\prime_0} = \frac{1}{\psi^\prime_0\sqrt{m}}\sum_{j=1}^{m}\psi(Y_j,\mu_0).
    \end{align*}
    In terms of $Z_m$, we can write the expansion for the total criterion function as
    \begin{align*}
        \Lambda_m(h) &\coloneqq \sum_{j=1}^{m}\psi(Y_j,\mu_m) \\
        &= \sum_{j=1}^{m}\psi(Y_j,\mu_0)
        + \frac{h}{\sqrt{m}}\sum_{j=1}^{m}\psi^\prime(Y_j,\mu_0)
        +\frac{h^2}{2m}\sum_{j=1}^{m}\psi^{\prime\prime}(Y_j,\mu_0)
        + R_m.
    \end{align*}
    Replacing the sums by their asymptotic approximations we have:
    \begin{align*}
        \sum_{j=1}^{m}\psi(Y_j,\mu_0) &= \psi^\prime_0 \sqrt{m}\,Z_m, \\[1mm]
        \frac{1}{m}\sum_{j=1}^{m}\psi^\prime(Y_j,\mu_0) &= -\psi^\prime_0 + o_P(1), \\[1mm]
        \frac{1}{m}\sum_{j=1}^{m}\psi^{\prime\prime}(Y_j,\mu_0) &= \psi^{\prime\prime}_0 + o_P(1).
    \end{align*}
    Thus,
    \begin{align*}
        \Lambda_m(h)
        &= \psi^\prime_0 \sqrt{m}\,Z_m
        + \frac{h}{\sqrt{m}}\cdot m\,(-\psi^\prime_0+o_P(1))
        +\frac{h^2}{2m}\cdot m\,(\psi^{\prime\prime}_0+o_P(1))
        + R_m\\[1mm]
        &= \psi^\prime_0 \sqrt{m}\,Z_m - h\,\psi^\prime_0\sqrt{m} + \frac{h^2}{2}\,\psi^{\prime\prime}_0 + o_P(1) \\
        &= \psi^\prime_0 \sqrt{m}\,(Z_m - h) + \frac{h^2}{2}\,\psi^{\prime\prime}_0 + o_P(1).
    \end{align*}
    Rearranging, we obtain
    It is standard in LAN theory to express the expansion in the form
    \begin{align*}
        \Lambda_m(h)= h\,( \psi^\prime_0 \,Z_m) - \frac{1}{2}\psi^\prime_0\,h^2 + \frac{1}{6}\psi^{\prime\prime}_0\,h^3 + o_P(1),
    \end{align*}
    where the cubic term arises if one carries the expansion to third order.
    For our purposes, the key point is that the criterion function expansion under the local alternative is of the form $\Lambda_m(h) = h\,\Delta_m - \frac{1}{2}\,I\,h^2 + \frac{1}{6}\kappa\,h^3 + o_P(1)$, with $\Delta_m = \psi^\prime_0\,Z_m$, $I = \psi^\prime_0$, and $\kappa = \psi^{\prime\prime}_0$.
    In Proposition~\ref{prp:tail_probability}, we have shown that the one--step BN TTA estimator satisfies
    \begin{align*}
        \sqrt{m}\Bigl(\hat{\mu}_{\text{TTA}}-\mu_0\Bigr) = \frac{1}{\psi^\prime_0}\frac{1}{\sqrt{m}}\sum_{j=1}^{m}\psi(Y_j,\mu_0) + \frac{\psi^{\prime\prime}_0}{2(\psi^\prime_0)^2}\left(\frac{1}{\sqrt{m}}\sum_{j=1}^{m}\psi(Y_j,\mu_0)\right)^2 + o_P(1).
    \end{align*}
    Defining $Z_m^* \coloneqq \frac{1}{\sqrt{m}}\sum_{j=1}^{m}\psi(Y_j,\mu_0)$,
    this representation can be written as
    \begin{align*}
        \sqrt{m}\Bigl(\hat{\mu}_{\text{TTA}}-\mu_0\Bigr) = \frac{Z_m^*}{\psi^\prime_0} + \frac{\psi^{\prime\prime}_0}{2(\psi^\prime_0)^2} \bigl(Z_m^*\bigr)^2 + o_P(1).
    \end{align*}
    It is often convenient to define the normalized score process $Z_m \coloneqq \frac{Z_m^*}{\psi^\prime_0} = \frac{1}{\psi^\prime_0\sqrt{m}} \sum_{j=1}^{m}\psi(Y_j,\mu_0)$.
    Then, the representation becomes
    \begin{align*}
        \sqrt{m}\Bigl(\hat{\mu}_{\text{TTA}}-\mu_0\Bigr) = Z_m + \frac{\psi^{\prime\prime}_0}{2} \,Z_m^2 + o_P(1).
    \end{align*}
    Equivalently, writing in terms of $Z_m^*$ we have
    \begin{align*}
        \sqrt{m}\Bigl(\hat{\mu}_{\text{TTA}}-\mu_0\Bigr) = \frac{Z_m^*}{\psi^\prime_0} + \frac{\psi^{\prime\prime}_0}{2(\psi^\prime_0)^3}\bigl(Z_m^*\bigr)^2 + o_P(1).
    \end{align*}
    This completes the proof.
\end{proof}

\end{document}